\documentclass[letterpaper, 10 pt, conference]{ieeeconf}  

\IEEEoverridecommandlockouts                              

\usepackage{graphics} 
\usepackage{epsfig} 
\usepackage{amsmath} 
\usepackage{amssymb}  
\usepackage{xcolor}
\usepackage{url}
\usepackage{booktabs}
\usepackage{caption}
\usepackage{subcaption}

\newcommand{\hide}[1]{}
\title{\LARGE \bf
Learning Agile Bipedal Motions on a Quadrupedal Robot
}

\author{Yunfei Li$^{1}$, Jinhan Li$^{1}$, Wei Fu$^{1}$ and Yi Wu$^{1,2}$
\thanks{$^{1}$Institute for Interdisciplinary Information Sciences, Tsinghua University, Beijing, China.
        {\tt\small liyf20@mails.tsinghua.edu.cn, jxwuyi@gmail.com}}%
\thanks{$^{2}$Shanghai Qi Zhi Institute, Shanghai, China.
        }%
}

\begin{document}

\maketitle
\thispagestyle{empty}
\pagestyle{empty}

\begin{abstract}
    Can a quadrupedal robot perform bipedal motions like humans? Although developing human-like behaviors is more often studied on costly bipedal robot platforms, we present a solution over a lightweight quadrupedal robot that unlocks the agility of the quadruped in an upright standing pose and is capable of a variety of human-like motions. Our framework is with a hierarchical structure. At the low level is a motion-conditioned control policy that allows the quadrupedal robot to track desired base and front limb movements while balancing on two hind feet. The policy is commanded by a high-level motion generator that gives trajectories of parameterized human-like motions to the robot from multiple modalities of human input. We for the first time demonstrate various bipedal motions on a quadrupedal robot, and showcase interesting human-robot interaction modes including mimicking human videos, following natural language instructions, and physical interaction. The video is available at \url{https://sites.google.com/view/bipedal-motions-quadruped}.
\end{abstract}

\section{Introduction}
Empowering robots with versatile motions like humans has been an important research topic to allow them to better coexist and interact with humans~\cite{hirai1998development}. 
Developing bipedal robot systems has attracted much interest since they have an appealing potential to mimic human behaviors thanks to their structural similarity to human beings~\cite{shamsuddin2011humanoid}. However, existing bipedal robots are typically expensive, heavy, and power-consuming~\cite{kuindersma2016optimization,shigemi2018asimo,stasse2017talos}. In contrast, quadrupedal robots are much cheaper and more lightweight and have recently demonstrated impressive sporting capabilities in various domains
~\cite{lee2020learning,miki2022learning,ji2022hierarchical}. This naturally raises an interesting question: \textit{Is it possible for quadrupedal robots to demonstrate agile human-like motions as an affordable alternative 
of humanoid robots?}


Enabling a quadrupedal robot to perform agile bipedal motions poses significant control challenges. Since quadrupedal robots are designed for dog-like behaviors with four legs on the ground, they must first stand upright on two feet from a four-leg pose at rest to unlock the motions of bipedal creatures. The stand-up procedure requires an agile control policy to gain enough momentum to swing up the robot and avoid flipping over at the same time. Furthermore, the bipedal motions are inherently unstable over common quadrupedal robots with spherical feet, thus the robot must actively adjust all its body parts to stay balanced once it stands up. Previous work utilized external mechanical support for the robot to stand up~\cite{yu2022multimodal}, 
while we aim to control a quadrupedal robot to mimic bipedal motions without any external hardware.

Another notable challenge is how to master a wide range of human-like motions. Due to the difference in kinematics and dynamics between humans and quadrupedal robots, it is difficult to directly track human motion capture data while keeping the quadruped robot balanced~\cite{xu2023language}. Therefore, motion encodings that can both represent versatile human-like behaviors and are also feasible for the embodiment of a quadrupedal robot require careful design.  

In this work, we present a hierarchical framework that enables agile bipedal motions on a quadrupedal robot. At the low level, we train a motion-conditioned policy with model-free reinforcement learning (RL) that is capable of balancing the quadrupedal robot on its two toes while tracking reference motions at the same time. We represent motions as a sequence of the desired state of the robot base and end effectors of the front limbs, which is flexible enough to encode a spectrum of behaviors and is plausible for the quadrupedal robot to execute. The policy is trained in a calibrated simulator and then transferred to the real robot. At the high level, a motion generator parameterizes human-like motions from videos or natural language descriptions into a sequence of desired targets for the low-level policy. 

\begin{figure}
    \centering
    \includegraphics[width=\linewidth]{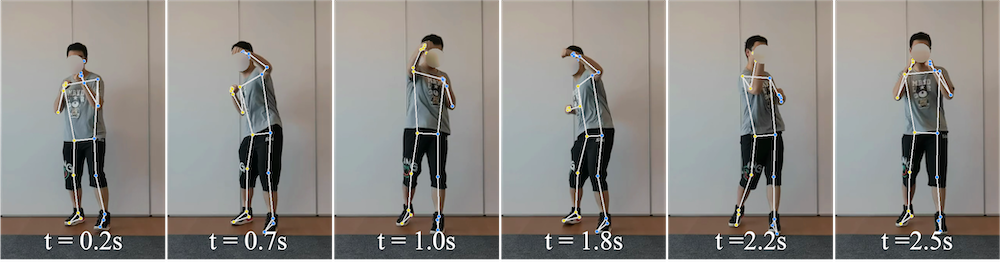}\vspace{1mm}
    \includegraphics[width=\linewidth]{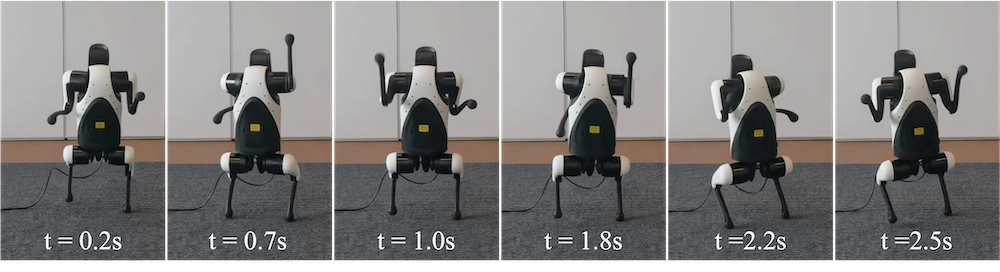}
    \caption{A quadrupedal robot demonstrates human-like motions with only hind feet on the ground. The top row shows the reference human boxing video. The bottom row shows the robot mimicking the human motion to perform multiple punches and uppercuts at a high speed.}
    \label{fig:expr:dance-boxing}
    \vspace{-6mm}
\end{figure}

We demonstrate the whole framework on an affordable quadruped platform Xiaomi CyberDog2 ($\sim$\$1800)~\cite{cyberdog2} and showcase the successful deployment of a variety of agile bipedal maneuvers and interactions with humans such as mimicking human videos to practice boxing (Fig.~\ref{fig:expr:dance-boxing}) and ballet dance, greeting commanded by natural language instructions, and walking hand-in-hand with a human. 
To the best of our knowledge, these agile bipedal motions are made possible on a quadrupedal robot for the first time.

\section{Related Work}

\textbf{Learning agile skills with quadrupedal robots:}
There has been tremendous progress in training a variety of agile skills on quadrupedal robot platforms with reinforcement learning, such as jumping over obstacles~\cite{bellegarda2020robust,margolis2022learning, park2021jumping,park2015online,smith2023learning}, landing~\cite{rudin2021cat}, soccer shooting~\cite{ji2022hierarchical}, and goalkeeping~\cite{huang2022creating}, but the motions are mainly with four legs on the ground. 
Only a few works study the possibility of bipedal motions on quadrupedal robots~\cite{fuchioka2023opt,smith2023learning}. Besides standing up and locomotion, we consider a wider range of motions that involve the base and hand movements to make more interesting interactions with humans. \cite{vollenweider2023advanced} trained a wheeled-legged robot to stand up and navigate and \cite{yu2022multimodal} used an external mechanical support to stand up, while we work on a canonical quadrupedal robot without special hardware.

As for learning robot motions, one popular line of work is motion imitation from reference trajectories. These works either directly mimic the reference motions~\cite{2018-TOG-SFV,RoboImitationPeng20} or adopt an adversarial approach~\cite{2021-TOG-AMP} to produce motions with the same style as the source dataset~\cite{li2023learning, 9981973, 10160751}.
Another research direction is to smartly parameterize motions of interest and track them with reward engineering~\cite{ji2022hierarchical,huang2022creating,margolis2023walk}.
Since the reference trajectories of bipedal motions for quadrupeds are not readily available, we choose to 
parameterize the motions using the base and front limb movements, which is versatile enough to represent a broad range of bipedal motions. 

\textbf{Developing controllers for bipedal motions:}
Traditional model-based methods like model predictive control~\cite{daneshmand2021variable,dantec2022whole} and trajectory optimization~\cite{apgar2018fast,hereid2019rapid} can obtain bipedal walking controllers that follow predefined gaits. They require accurate modeling of the robot dynamics and state estimation, and run intensive optimization online to achieve good performances.  
Model-free reinforcement learning (RL) is another direction to obtain such controllers that alleviates the burden of heavy engineering in dynamics modeling and has demonstrated superior performance in bipedal velocity following~\cite{li2021reinforcement} and jumping~\cite{li2023robust}.
In this work, we develop a model-free RL method to learn bipedal skills, but over an affordable \textit{quadrupedal robot} that is not specialized for bipedal motions. We also work beyond locomotion tasks~\cite{cheng2023extreme,go2} and demonstrate more versatile motions after freeing the robot's front limbs from walking. 

\textbf{Sim-to-real transfer:} 
Domain randomization is a powerful technique to bridge the sim-to-real gap when deploying a policy trained in simulation to a real robot~\cite{tobin2017domain,peng2018sim,mehta2020active}. The applied randomization range typically requires expertise to design~\cite{vuong2019pick}. System identification using real data is another direction~\cite{ljung1998system,kolev2015physically}, such as learning a motor model to fit its complex dynamics~\cite{hwangbo2019learning}. There are also works that iteratively calibrate the simulator using learned trajectories and optimize the policy with new simulation parameters, and demonstrate successful transfer in precise robot arm manipulation~\cite{chebotar2019closing} and bipedal motions~\cite{tan2016simulation}.
We similarly leverage some real-world data to tune the randomization range of critical parameters in simulation to reduce the discrepancy between simulation and the real world for a successful policy transfer.

\section{Preliminary}

\hide{We parameterize the quadrupedal robot's motions using two components: the base velocities and the positions of the front hands in the robot base frame.} 
We formulate the bipedal motion learning as a Markov Decision Process (MDP) defined by $(\mathcal{S},\mathcal{A},\mathcal{T},r,\gamma, \rho_0)$, where $\mathcal{S}$ is the state space, $\mathcal{A}$ is the action space, $\mathcal{T}:\mathcal{S}\times \mathcal{A} \mapsto \mathcal{S}$ is the transition function, $r:\mathcal{S}\times \mathcal{A} \mapsto \mathbb{R}$ is the reward function, $\gamma$ is the discount factor, and $\rho_0$ is the initial state distribution. The objective is to train a policy $\pi^*$ which maximizes the discounted accumulative reward $\pi^*=\arg\max_{\pi}\mathbb{E}_{s_0\sim \rho_0, a_t\sim \pi(\cdot|s_t)} \left[ \sum_{t\geq 0} \gamma^t r(s_t,a_t)\right]$.
We adopt proximal policy optimization (PPO)~\cite{schulman2017proximal}, a state-of-the-art RL algorithm to solve the MDP. PPO jointly trains a policy network $\pi_{\theta}(a|s)$ and a critic function $V_{\phi}$\hide{ when optimizing the RL objective}.

We consider bipedal motions with an upright standing pose in this work, and parameterize them with the linear velocity of the robot base, the base heading, and the positions of the end effectors in front limbs relative to the base.

\section{Method}
We propose a two-level framework to learn bipedal motions on a quadruped. We first train a motion-conditioned policy in simulation that allows the robot to stand on hind toes while tracking random motions. Since the desired bipedal motions are highly agile and are sensitive to physical parameters, we calibrate the simulator via a simple real-to-sim process to enable successful deployment on the real robot. Afterwards, we generate the sequence of motion targets from multiple modalities of human inputs, and command the RL policy to accomplish human-like agile bipedal maneuvers.

\subsection{Learning a motion-conditioned policy with RL}

We design a model-free RL approach to obtain a control policy that empowers a quadrupedal robot with the ability to stand up and track motions. The policy is trained in a massively parallel GPU-based simulator Isaac Gym~\cite{makoviychuk2021isaac}. 

\textbf{Observations and actions:} Our policy observes a history of proprioceptive information as its input and predicts the PD control target for all 12 motors. Specifically, the observation concatenates 3 frames of sensory input at $[t - 0.04s, t - 0.02s, t]$. Each frame consists of joint positions, the orientation of the robot base, the last applied actions, the desired linear and angular velocity of the robot base, and the desired positions of front toes in the base coordination. 
We encode the base orientation using the projection of the vectors $(0, 0, -1)$ and $(1, 0, 0)$ from the world coordination to the current robot base coordination. The policy predicts the target joint positions for the PD controller with parameters $K_p=30$ and $K_d=3$ at 50Hz. The critic function takes the policy observation and other privileged information that is only accessible in simulation such as the joint friction and damping as its input.


\textbf{Reward design:} The reward function is a summation of three categories of terms $r = r^{\textrm{stand}} + r^{\textrm{track}} + r^{\textrm{reg}}$ to achieve the following objectives: maintaining the standing pose with only two toes contacting the ground, tracking desired motions as accurately as possible, and avoiding drastic behaviors that are dangerous when deployed in the real world. 

We define the standing reward similar to \cite{smith2023learning} as $r^{\textrm{stand}}=r^{\textrm{height}} + r^{\textrm{pitch}} + r^{\textrm{collision}}$, where $r^{\textrm{height}}$ encourages the robot to lift up its body, $r^{\textrm{pitch}}$ credits the robot to maintain a certain pitch angle so as to stand upright, and $r^{\textrm{collision}}$ penalizes all parts of the robot except its rear feet touching the ground. 

The tracking reward $r^{\textrm{track}}$ aims to match the linear velocity, the heading of the robot base, and the relative position of the front toes to their targets.
$r^{\textrm{track}}(s) = r^{\textrm{track}}_{\textrm{base\_v}}(s) + r^{\textrm{track}}_{\textrm{heading}}(s) + r^{\textrm{track}}_{\textrm{hand}}(s)$, $r^{\textrm{track}}_{x}(s) = \alpha_{x} c_{x}(s) \exp\left(-\frac{e_x(s)}{\sigma_x}\right), x=\{\textrm{base\_v}, \textrm{heading}, \textrm{hand}\}$.
$\alpha_x$ and $\sigma_x$ are weighting constants, $e_x(s)$ is the tracking error. $c_x(s)$ is a dynamic scaling factor in the range of $[0, 1]$ conditioned on the standing performance and only reaches 1 after the robot has stood upright. 
If $c$ is a constant, we find it challenging to obtain a policy that both stands up high and tracks motion since reducing tracking error is easier to exploit compared to the standing reward. We observe that the robot would get stuck at a strategy that tracks hand motions while sitting on the hind legs (see Sec.~\ref{sec:expr:ablate}).

The regularization reward $r^{\textrm{reg}}$ sums up necessary shaping terms that penalize unrealistic behaviors. $r^{\textrm{reg}}$ includes commonly used terms from ~\cite{rudin2022learning} that penalize drastic joint motions, joint positions and torques close to limits, and large action rates. In addition, we regulate rear foot movements to follow a trotting gait with a height of 5cm with $r^{\textrm{reg}}_{\textrm{gait}}$ and penalize slipping on the ground with $r^{\textrm{reg}}_{\textrm{slip}}$.


\textbf{Episode termination:}
The robot resets from a sitting pose with four feet touching the ground. An episode terminates after a maximum number of 1000 steps is reached or under any of the following early-termination conditions: (a) the robot base or the front limbs is in collision after the first 30 steps, (b) any joint reaches its position limit. 

\textbf{Tracking targets:}
The agent is commanded to track random motion targets during RL training.
The desired base linear velocity along the $x$-axis (forward and backward) is sampled from discretized bins ranging between [-0.3, 0.3]m/s and discretized at an interval of 0.1m/s every 10 seconds. The desired velocity in the $y$-axis (side) is fixed as 0. The desired heading direction is sampled from [$-\pi/2$, $\pi/2$] radians relative to the current heading every 10 seconds, and the desired angular velocity (which is the observation of the policy) is updated every step using the difference between the current and desired heading directions.
We update future goal positions of front limb end-effectors every 3 seconds, and compute the desired positions of front toes in each step (the observed targets) by interpolating between the last and future goal positions linearly. 
We make sure that the goal positions of end-effectors are reachable within the joint limit. 

\textbf{A sit-down policy for safe ending:} We train a separate sit-down policy that can transit the robot from random stand-up poses to a quadrupedal landing pose, and append it after the motion-conditioned policy for safe termination during deployment on the real robot. The sit-down is also trained with RL. The initial state distribution for the policy is upright standing poses with random facing directions and random front limb motor positions sampled within their limits. The reward function is designed to encourage the robot's belly to face downward and to credit the joint positions for being close to those in a nominal quadrupedal standing pose.

\subsection{Sim-to-real transfer via real-to-sim calibration}\label{sec:method:real2sim}


Domain randomization is a powerful sim-to-real technique and we also adopt it for deployment. However, bipedal motions are inherently unstable and are sensitive to physical parameters. The RL policy would fail to find a solution that fits all physical parameters (see Sec.~\ref{sec:expr:ablate}) if too much randomization is applied. To ensure good transfer performance with only slight randomization, we conduct real-to-sim calibration that searches for the simulation parameters that can best explain the real robot trajectories. 

Specifically, we apply the same sequence of actions both in simulation and on the real robot for the calibration. The sequence used to probe the real world is from rollouts of policies trained in the early development stage of this project and lasts for 120 seconds in total. 
Since the policies trained from an uncalibrated simulator can hardly succeed in the real world, we choose to hang up the robot and run the sequence in an open loop during calibration to avoid hardware damage, and save the joint position readings $\textbf{q}^{\textrm{real}}_i|_{i=0}^{N}$ from 12 motors at 200 Hz. Given a specific configuration of simulation parameters $\xi$, we fix the base link in simulation and collect joint positions $\textbf{q}^{\textrm{sim}}_i(\xi)|_{i=0}^{N}$ by applying the same action sequence. We spawn 8192 simulation environments and search for the parameters with minimal discrepancy to the real world as
\begin{equation}
    \xi^\star = \arg\min_{\xi} \sum_{i=0}^{N}|\textbf{q}^{\textrm{sim}}_i(\xi) - \textbf{q}^{\textrm{real}}_i|^2.
\end{equation}
We calibrate joint friction, joint damping, limb mass, and system delay in this process. To avoid overfitting to the action sequence used in calibration, we add margins to the best sampled values.
Other parameters that are not probed (e.g., those related to contacts) are randomized naively. 
All randomized physical parameters and their ranges are listed in Table~\ref{tab:domain_randomization}.
We remark that our calibration process is more end-to-end and does not require additional instruments compared to fitting a single module such as motors in~\cite{hwangbo2019learning}.
\begin{table}
    \caption{Randomized simulation parameters and ranges.}
    \vspace{-1mm}
    \label{tab:domain_randomization}
    \centering
    \begin{tabular}{ll}
        \toprule
        Name & Range \\
        \midrule
        Joint friction & [0.03, 0.08] \\
        Joint damping & [0.02, 0.06] \\
        Rigid body friction & [1.0, 3.0] \\
        Rigid body restitution & [0.0, 0.4] \\
        Base mass offset & [-0.5kg, 0.5kg] \\
        Hip mass offset & [0kg, 0.1kg] \\
        Thigh mass offset & [-0.05kg, 0.05kg] \\
        Calf mass offset & [-0.05kg, 0.05kg] \\
        Foot mass offset & [0kg, 0.01kg] \\
        Center of mass displacement & [-0.01m, 0.01m] \\
        $K_p$, $K_d$ & [80\%, 120\%] \\
        Delay & [0.005s, 0.03s] \\
        \bottomrule
    \end{tabular}
    \vspace{-3mm}
\end{table}


\subsection{Generating reference motions for human-like maneuvers}\label{sec:method:motion_gen}
\begin{figure}
    \centering
    \includegraphics[width=0.98\linewidth]{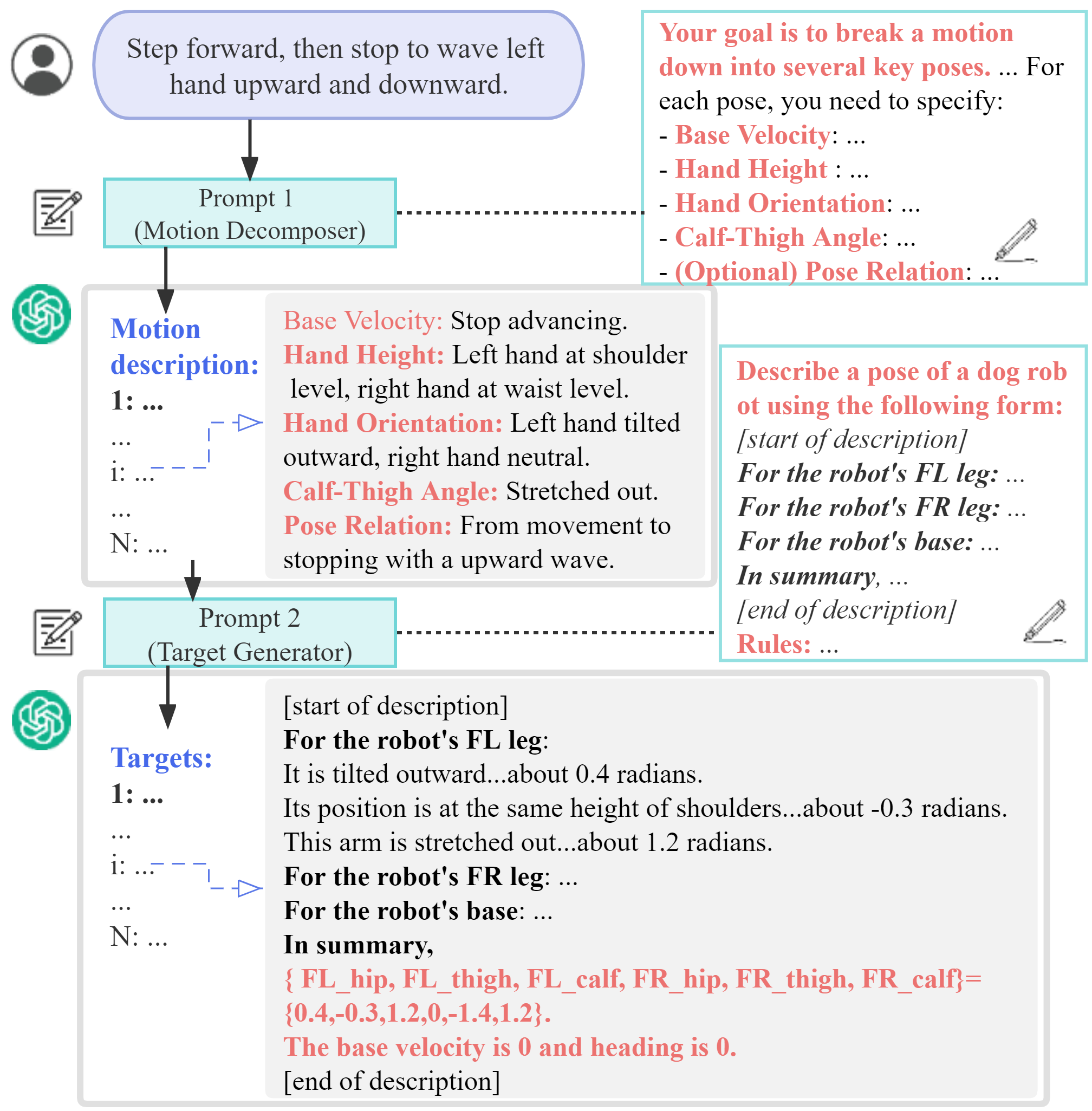}
    \caption{The workflow of generating reference motions from human language instructions with an LLM. The language command from the user is first decomposed into a sequence of motion descriptions, then converted to targets consisting of base velocity, heading, and front limb joint positions. The example outputs by the LLM in both steps are in grey boxes, and the prompts we use are in cyan boxes.}
    \label{fig:method:llm-prompts}
    \vspace{-6mm}
\end{figure}

We study the generation of human-like motions for a quadrupedal robot from two modalities: one is to mimic human videos, and another is to convert natural language instructions into motions with a large language model (LLM).

For mimicking human videos, we focus on retargeting the human front limb motions to the robot. We first adopt an off-the-shelf human pose detector~\cite{bazarevsky2020blazepose,xu2020ghum} to estimate the 3D skeleton and obtain the vectors of wrists relative to the body $p_{\textrm{human}}$ for each frame extracted from the video. The desired hand positions for the robot $p_{\textrm{robot}}$ are then calculated by scaling down $p_{\textrm{human}}$ to compensate for the differences in the size and working space between the human and the robot. 


As for the language input, we leverage the common sense knowledge of a pretrained LLM to generate reference motions that can fulfill the natural language instruction. As illustrated in Fig.~\ref{fig:method:llm-prompts}, the generation is done with two rounds of conversation. In the first round, we prompt the LLM to decompose an abstract instruction into a sequence of key frames and to give detailed descriptions for each frame along the axis of the base velocity, hand height, hand orientation, calf-thigh joint, and the relationship with the previous frame. In the second round, the LLM is prompted to format the description of each frame as the precise target motion for the robot, including the base velocity, the heading direction, and the joint positions of 6 motors on the front limbs. The joint positions are finally converted to positions of end effectors with forward kinematics.

Since the LLM has limited domain knowledge concerning the specification of our quadrupedal robot, it is challenging to directly generate precise desired motions. To help the LLM generate reasonable values, we provide it with a set of rules specific to the kinematics of the robot in the prompt similar to \cite{yu2023language,tang2023saytap}, such as ``if the FL hand is tilted outward, then set FL\_hip\_joint within range (0.1,0.57) radians''. 


\section{Experiments}
We conduct experiments with a quadrupedal robot Xiaomi CyberDog2~\cite{cyberdog2}. The robot is driven by 12 identical CyberGear~\cite{cybergear} motors. All computation in deployment is run on the internal board of the robot to reduce the communication latency.
\subsection{Performance of the motion-conditioned policy}
\begin{figure*}
    \centering
    \includegraphics[width=\linewidth]{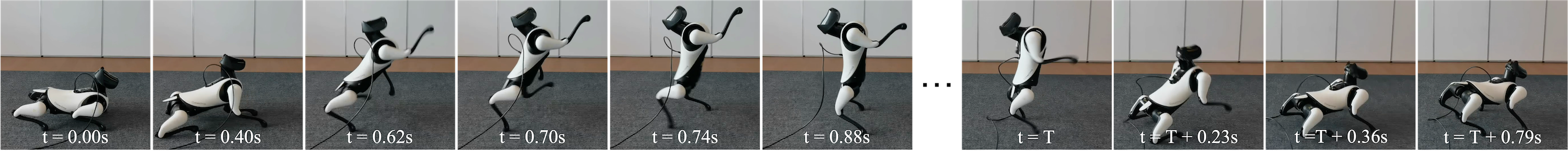}
    \caption{Our RL policy brings up the quadrupedal robot from a lying pose to a stabilized bipedal standing pose. The separate sit-down policy then controls the robot from the upright standing pose to settle down with four legs on the ground. The learned policies demonstrate great agility, using less than 1 second to stand up and sit down, while are sufficiently robust to work on the real robot.}
    \label{fig:expr:standup}
    \vspace{-2mm}
\end{figure*}
\begin{figure*}
    \centering
    \begin{subfigure}[b]{0.48\linewidth}
        \centering
        \includegraphics[width=0.49\linewidth]{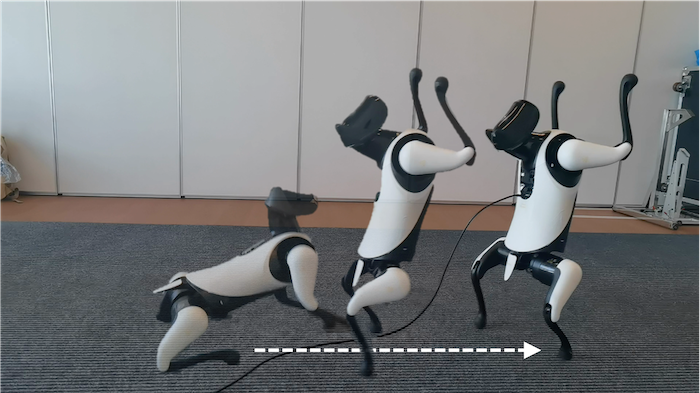}
        \includegraphics[width=0.49\linewidth]{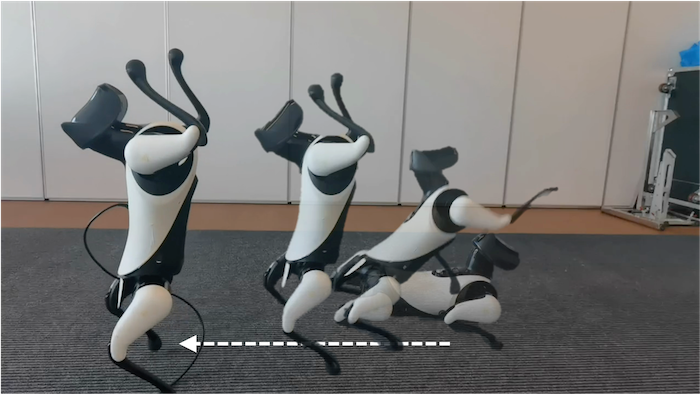}
        \caption{Track different linear velocities of the robot base.}\label{fig:expr:loco-linv}
    \end{subfigure}
    \hspace{2mm}
    \begin{subfigure}[b]{0.48\linewidth}
        \centering
        \includegraphics[width=0.49\linewidth]{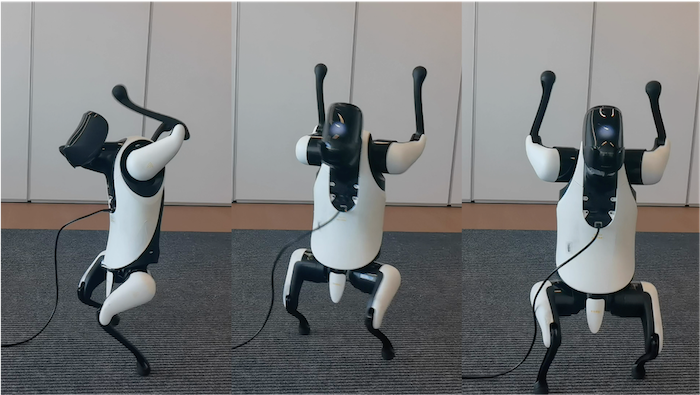}
        \includegraphics[width=0.49\linewidth]{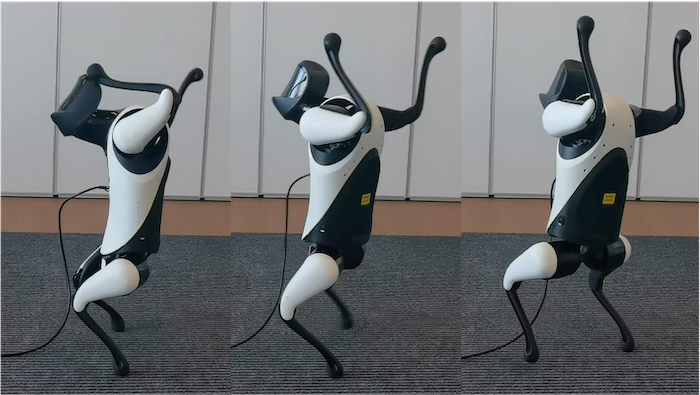}
        \caption{Track different heading directions of the robot base.}
        \label{fig:expr:loco-ang}
    \end{subfigure}
    \caption{The quadrupedal robot demonstrates bipedal locomotion following target linear velocities $v_x=\pm0.3\textrm{m/s}$ to walk forward or backward, and tracking target heading directions 90 degrees to the left or 60 degrees to the right.}
\end{figure*}
    
    
\begin{table*}
    \caption{Ablation studies on the key designs for training our RL policy. The mean and standard deviation over three seeds are reported for each variant. Real2sim calibration, the dynamic scaling factor in the tracking reward, and the feet regularization reward all contribute to achieving the best balance over the stand-up performance, the tracking accuracy, and the feet clearance. }
    \label{tab:expr:ablation}
    \centering
    \begin{tabular}{lccccccc}
    \toprule
        Method & $r^{\textrm{track}}_{\textrm{base\_v}}$ & $r^{\textrm{track}}_{\textrm{heading}}$ & $r^{\textrm{track}}_{\textrm{hand}}$ & $r^{\textrm{height}}$ & $r^{\textrm{reg}}_{\textrm{slip}}$ & Episodic reward & Episode length\\
    \midrule
        Ours & \textbf{0.221$\pm$0.022} & \textbf{0.173$\pm$0.009} & \textbf{0.679$\pm$0.010} & \textbf{0.389$\pm$0.005} & \textbf{-0.122$\pm$0.005} & \textbf{30.53$\pm$0.58} & \textbf{826.68$\pm$12.01}\\
        w/o real2sim & 0.080$\pm$0.013 & 0.074$\pm$0.013 & 0.318$\pm$0.059 & 0.194$\pm$0.036 & -0.128$\pm$0.020 & 12.10$\pm$2.35 & 413.14$\pm$73.35\\
        w/o dynamic scale & 0.177$\pm$0.058 & 0.137$\pm$0.050 & 0.378$\pm$0.327 & 0.300$\pm$0.098 & -0.126$\pm$0.027 & 21.92$\pm$9.43 & 767.41$\pm$106.08\\
        w/o feet reg. & 0.211$\pm$0.040 & 0.153$\pm$0.007 & 0.612$\pm$0.012 & 0.350$\pm$0.014 & -0.207$\pm$0.076 & 23.82$\pm$1.59 & 749.31$\pm$20.00\\
        \bottomrule
    \end{tabular}
    \vspace{-3mm}
\end{table*}

We test whether the control policy trained with RL can accomplish bipedal locomotion tasks with manually written desired motions. When commanded to track zero velocity and zero relative heading direction, the robot stands upright from the initial lying pose within 1 second as illustrated in the left half of Fig.~\ref{fig:expr:standup}. Fig.~\ref{fig:expr:loco-linv} shows the robot stably walks forward and backward following the target linear velocity $v_x=\pm 0.3$m/s after standing up. In Fig.~\ref{fig:expr:loco-ang}, we set the desired heading direction to 90 and -60 degrees relative to the initial pose, and the policy effectively controls the robot to turn around following these commands. After each bipedal motion, the robot is controlled by the same sit-down RL policy to settle down back to a resting pose with four legs on the ground (see the right half of Fig.~\ref{fig:expr:standup}).  

\begin{figure}
    \centering
    \includegraphics[width=0.9\linewidth]{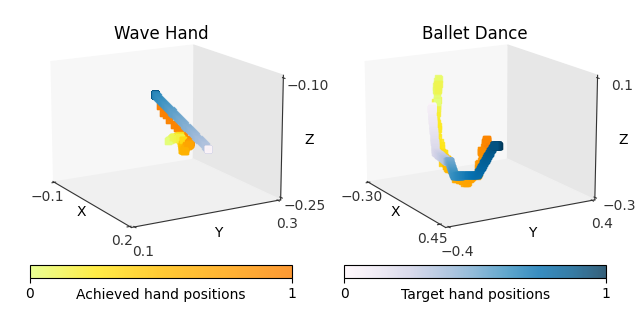}
    \vspace{-3mm}
    \caption{The visualization of front-toe tracking during segments of ``wave hand'' (left) and ``ballet dance'' (right) in simulation. Blue dots represent the desired positions and the orange dots represent the achieved positions. The varying hues indicate the progression of time. }
    \label{fig:expr:foot_traj}
    \vspace{-3mm}
\end{figure}

We visualize the tracking performance of front toes in simulation during segments of ``waving hand'' (left) and ``ballet dance'' (right) in Fig.~\ref{fig:expr:foot_traj}. The intended positions (blue) and the achieved positions (orange) are closely aligned, showcasing the effectiveness of front-toe tracking.

\textbf{Results of real-to-sim calibration:}
\begin{figure*}
    \centering
    \includegraphics[width=0.23\linewidth]{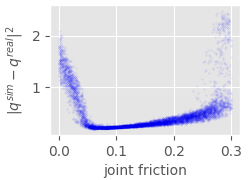}
    \includegraphics[width=0.7\linewidth]{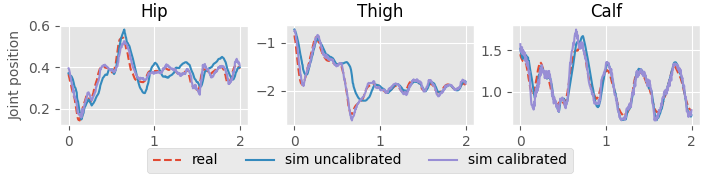}
    \caption{The results of real-to-sim calibration. Left: the relationship between the sim-to-real difference and the simulated joint friction. The calibrated randomization range is chosen in the region with low errors. Right: the real motor positions w.r.t. the time on the rear left leg and the simulated ones using uncalibrated and calibrated physical parameters. The simulated trajectory matches the real world better after real-to-sim calibration. }
    \label{fig:expr:calibration}
    \vspace{-2mm}
\end{figure*}
As is described in Sec.~\ref{sec:method:real2sim}, we conduct real-to-sim calibration to reduce the discrepancy between the simulator and the real robot. In the left plot of Fig.~\ref{fig:expr:calibration}, we show the relation between simulated joint frictions and the sim-to-real discrepancy in joint positions per control step averaged over 120s of the open-loop trajectory. The sim-to-real gap is the smallest in our calibrated joint friction range, and the error goes up in both directions out of this range. In the right plot, we visualize the positions of three motors on the rear left leg in simulation and the corresponding real robot data recorded during open-loop control. The trajectory generated from the best-calibrated physical parameters (purple) matches the real data (red) much better than the uncalibrated one (blue). 

\subsection{Performing human-like motions when combined with generated motion targets}
\begin{figure}
    \centering
    \includegraphics[width=\linewidth]{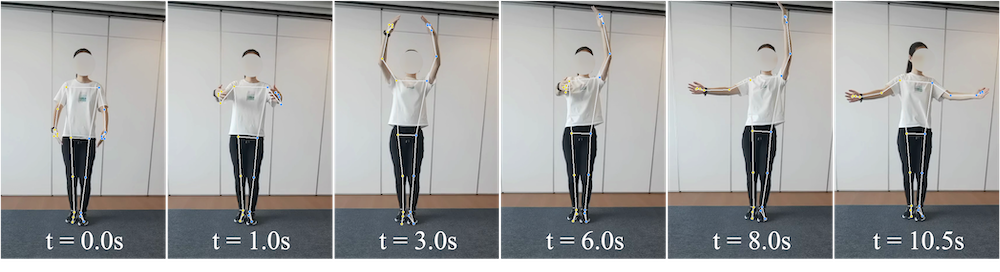}\vspace{1mm}
    \includegraphics[width=\linewidth]{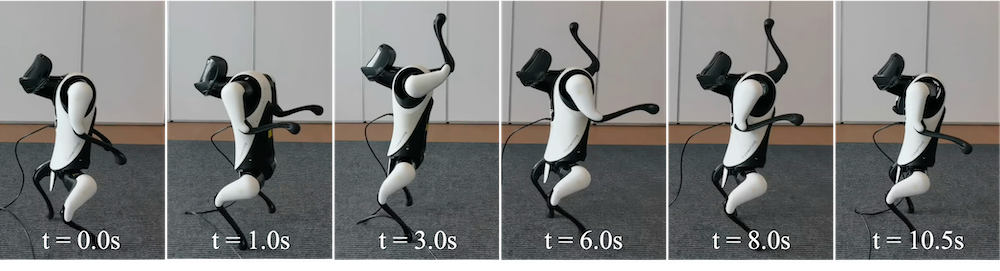}
    \caption{The quadruped robot follows a human to perform ballet. It keeps balanced and tracks the hand poses at best effort during the long motion that lasts for more than 10s.}
    \label{fig:expr:dance-ballet}
    \vspace{-3mm}
\end{figure}

We then command the low-level policy with target commands parsed from human videos or natural languages and verify whether our system can enable human-like motions on the quadruped robot. 
We invite two human participants to provide motion clips and map their hand trajectories to the quadruped as described in Sec.~\ref{sec:method:motion_gen}. Fig.~\ref{fig:expr:dance-boxing} demonstrates our robot mimicking fast boxing motion. The commands are extracted every 0.1s from the video. The robot manages to keep balance while performing punches and an uppercut at high speed. The human reference frames annotated with detected skeletons are illustrated in the front row, and the robot execution trajectories are in the bottom row. Fig.~\ref{fig:expr:dance-ballet} shows a slow ballet dance that lasts for more than 10 seconds. The motions are extracted every 0.5s. The robot follows the human guidance to gracefully lift up both hands from its waist to above its head, then drops down the right arm and opens it up with its best effort, and finally moves the left hand to the side. The robot does open its arm to the same extent as the human due to the position limit of its hip joints.

\begin{figure}
    \centering
    \includegraphics[width=0.9\linewidth]{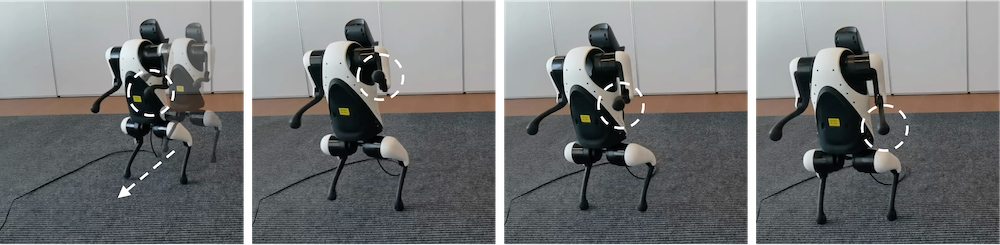}
    \caption{Using LLM to generate motions for walking forward a few steps and then waving the left hand up and down. }
    \label{fig:expr:llm-wavehand}
    \vspace{-5mm}
\end{figure}

We also try to generate motion sequences to fulfill bipedal motions with an LLM. We prompt GPT-3.5~\cite{ouyang2022training} to decompose a natural language instruction into a sequence of target base velocity, heading direction, and front limb joint positions, then convert joint positions to hand positions with forward kinematics. Fig.~\ref{fig:expr:llm-wavehand} shows an example of using LLM-generated commands to follow a language instruction ``step forward, then stop to wave left hand upward and downward''. 
\begin{figure}
    \centering
    \includegraphics[width=\linewidth]{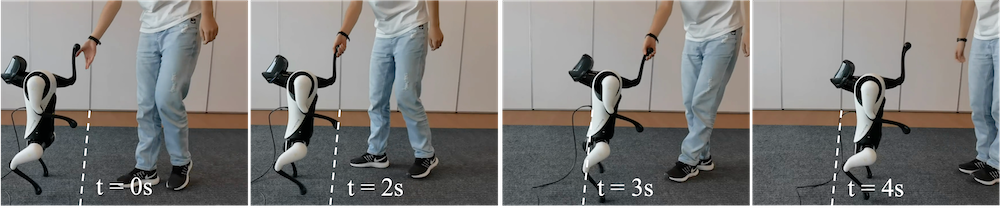}
    \caption{A human walks hand-in-hand with the standing quadruped robot and drags its front leg to change its position.}
    \label{fig:expr:human-interaction}
\end{figure}

Finally, we show an interesting example of physical interaction in Fig.~\ref{fig:expr:human-interaction} where a human holds the lifted front limb of the standing quadruped and physically guides its position. Our policy is sufficiently robust to keep balanced at all times.

\subsection{Ablation studies}\label{sec:expr:ablate}

We compare our method with the following variants: (a) w/o real2sim calibration, which randomizes the joint friction according to the value provided in the URDF to illustrate an uncalibrated simulation environment; (b) replacing the dynamic scales $c_x(s)$ in the tracking reward terms that are conditioned on the standing up performance with static coefficients 1; (c) removing the reward terms $r^{\textrm{reg}}_{\textrm{gait}}$ and $r^{\textrm{reg}}_{\textrm{slip}}$ that regulate the gait of rear feet. 
All variants are trained for three seeds. Each run is evaluated for 50 episodes in the same calibrated environment using the checkpoint trained for 18000 iterations. The performances are measured using both overall episodic metrics and detailed rewards regarding the base height, the tracking error, and the feet clearance.

\begin{figure}
    \centering
    \includegraphics[width=0.9\linewidth]{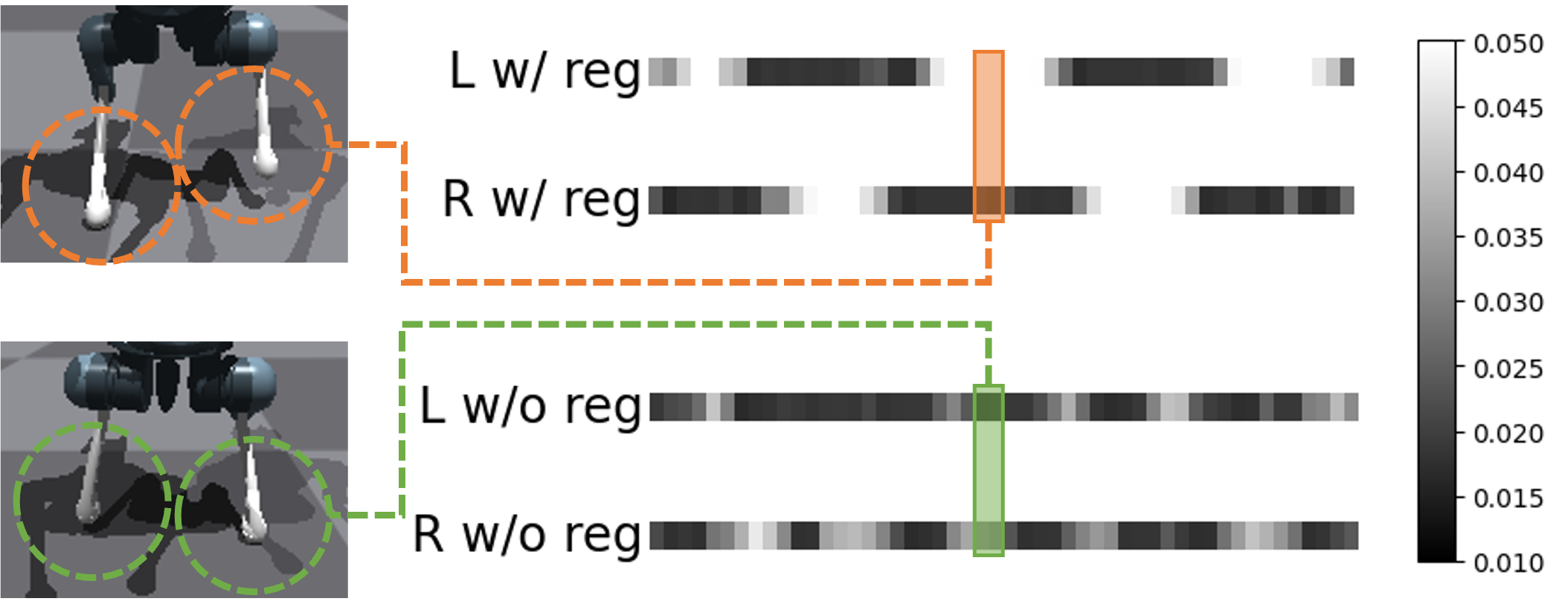}
    \caption{The feet heights recorded along the execution of different policies in simulation. Darker colors indicate the lower heights. Top: The left and right foot heights of the policy trained with our feet regularization reward terms, which demonstrates nice feet clearance. Bottom: A policy trained without these regularization terms tends to slip on the ground and can hardly be transferred to the real robot. }
    \label{fig:expr:feetclr}
    \vspace{-3mm}
\end{figure}

As is shown in Table~\ref{tab:expr:ablation}, using an uncalibrated joint friction range [0, 0.2] (the second row) significantly degrades the performance in all metrics compared with the calibrated range of [0.03, 0.08] (the first row), indicating the necessity of an appropriate randomization range. 
The variant ``w/o dynamic scale'' also performs worse than our original version. It results in a lower base height since the robot finds a sub-optimal strategy that ``sits'' on the hind calf. The tracking reward and the base height of the policy trained without the gait regularization are comparable to the main result, but the feet slippery metric is much worse. We also compare the foot heights of the policies trained with or without the gait regularization in Fig.~\ref{fig:expr:feetclr}. Without the regualization, the robot only lifts up the hind legs slightly above the ground and demonstrates irregular contact patterns, thus is difficult to be deployed in the real world where the materials of the ground and the feet are not perfectly rigid bodies. 

\section{Conclusion}
We study the problem of enabling a quadrupedal robot to perform agile human-like bipedal motions and propose a bi-level framework. The low level is a motion-conditioned RL policy that tracks the desired states of the robot base and the front limbs while balancing on hind toes. At the high level, we generate human-like motion sequences to command the low-level policy from human videos or natural language instructions. Currently, we consider motions that are feasible with proprioceptive states only. Augmenting the robot with environmental perception to perform more complex interactions with humans and objects is an interesting future work.

\section*{ACKNOWLEDGMENT}
Yi Wu is supported by 2030 Innovation Megaprojects of China (Programme on New Generation Artificial Intelligence) Grant No. 2021AAA0150000. We thank Yangwei You, Tianlin Liu, and other colleagues at Xiaomi for the support of robots and valuable discussions.

\bibliographystyle{IEEEtran}
\bibliography{reference}

\end{document}